\documentclass[10pt,twocolumn,letterpaper]{article}

\usepackage{wacv}
\usepackage{times}
\usepackage{epsfig}
\usepackage{graphicx}
\usepackage{amsmath}
\usepackage{amssymb}
\usepackage{verbatim}

\wacvfinalcopy 

\def\wacvPaperID{213} 

\begin{document}
\title{Interpretable Visual Question Answering \\ by Visual Grounding from Attention Supervision Mining}

\author{Yundong Zhang \\
Stanford University\\
{\tt\small ydzhang12345@gmail.com}
\and
Juan Carlos Niebles \\
Stanford University\\
{\tt\small jniebles@cs.stanford.edu}
\and
Alvaro Soto \\
Universidad Catolica de Chile\\
{\tt\small asoto@ing.puc.cl}
}

\maketitle
\ifwacvfinal\thispagestyle{empty}\fi

\begin{abstract}
A key aspect of VQA models that are interpretable is their ability to ground their answers to relevant regions in the image. Current approaches with this capability rely on supervised learning and human annotated groundings to train attention mechanisms inside the VQA architecture. Unfortunately, obtaining human annotations specific for visual grounding is difficult and expensive. In this work, we demonstrate that we can effectively train a VQA architecture with grounding supervision that can be automatically obtained from available region descriptions and object annotations. We also show that our model trained with this mined supervision generates visual groundings that achieve a higher correlation with respect to manually-annotated groundings, meanwhile achieving state-of-the-art VQA accuracy.
\end{abstract}

\section{Introduction}


We are interested in the problem of visual question answering (VQA), where an algorithm is presented with an image and a question that is formulated in natural language and relates to the contents of the image. The goal of this task is to get the algorithm to correctly answer the question. The VQA task has recently received significant attention from the computer vision community, in particular because obtaining high accuracies would presumably require precise understanding of both natural language as well as visual stimuli. In addition to serving as a milestone towards visual intelligence, there are practical applications such as development of tools for the visually impaired.

The problem of VQA is challenging due to the complex interplay between the language and visual modalities.
On one hand, VQA algorithms must be able to parse and interpret the input question, which is provided in natural language \cite{Geman:EtAl:2015,VQAFirst:EtAl:2014,vqa}. This may potentially involve understanding of nouns, verbs and other linguistic elements, as well as their visual significance. On the other hand, the algorithms must analyze the image to identify and recognize the visual elements relevant to the question. Furthermore, some questions may refer directly to the contents of the image, but may require external, common sense knowledge to be answered correctly.
Finally, the algorithms should generate a textual output in natural language that correctly answers the input visual question.
In spite of the recent research efforts to address these challenges, the problem remains largely unsolved \cite{VQA:Survey:2017}. 

\begin{figure}[t]
\begin{center}
\includegraphics[width=0.5\textwidth]{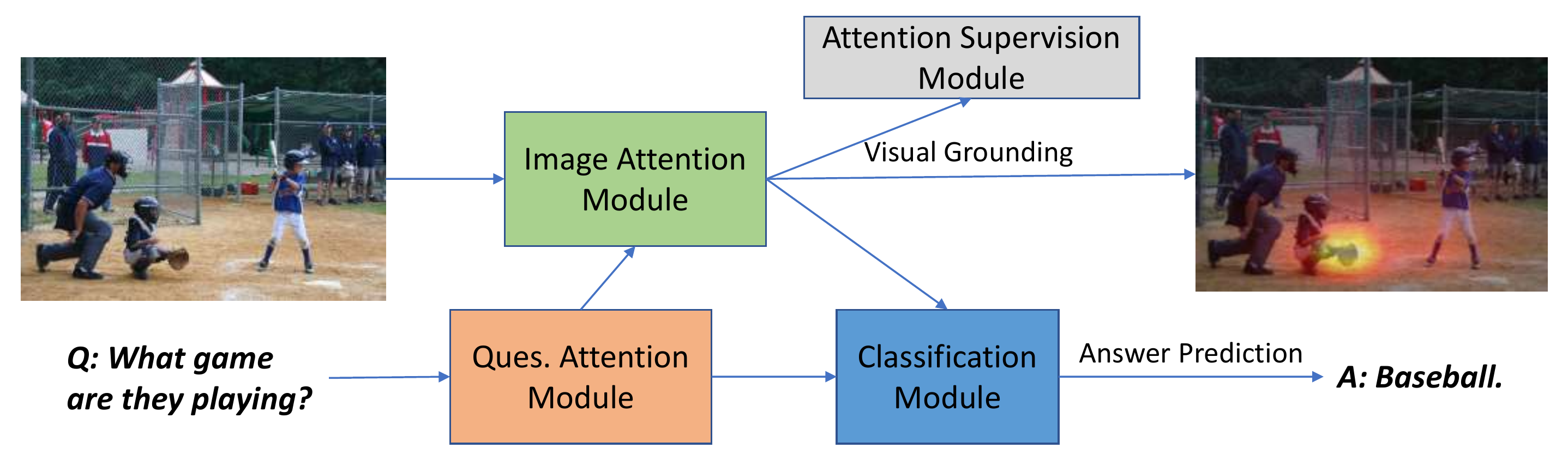}
\end{center}
\caption{Interpretable VQA algorithms must ground their answer into image regions that are relevant to the question. In this paper, we aim at providing this ability by leveraging existing region descriptions and object annotations to construct grounding supervision automatically.}
\end{figure}

We are particularly interested in giving VQA algorithms the ability to identify the visual elements that are relevant to the question. In the VQA literature, such ability has been implemented by \emph{attention} mechanisms. Such attention mechanisms generate a heatmap over the input image, which highlights the regions of the image that lead to the answer. These heatmaps are interpreted as groundings of the answer to the most relevant areas of the image. Generally, these mechanisms have either been considered as latent variables for which there is no supervision, or have been treated as output variables that receive direct supervision from human annotations. Unfortunately, both of these approaches have disadvantages. First, unsupervised training of attention tends to lead to models that cannot ground their decision in the image in a human interpretable manner. Second, supervised training of attention is difficult and expensive: human annotators may consider different regions to be relevant for the question at hand, which entails ambiguity and increased annotation cost. Our goal is to leverage the best of both worlds by providing VQA algorithms with interpretable grounding of their answers, without the need of direct and explicit manual annotation of attention.

From a practical point of view, as autonomous machines are increasingly finding real world applications, there is an increasing need to provide them with suitable capabilities to explain their decisions. However, in most applications, including VQA, current state-of-the-art techniques operate as black-box models that are usually trained using a discriminative approach. Similarly to \cite{vqa-hat}, in this work we show that, in the context of VQA, such approaches lead to internal representations that do not capture the underlying semantic relations between textual questions and visual information. Consequently, as we show in this work, current state-of-the-art approaches for VQA are not able to support their answers with a suitable interpretable representation.


In this work, we introduce a methodology that provides VQA algorithms with the ability to generate human interpretable attention maps which effectively ground the answer to the relevant image regions.
We accomplish this by leveraging region descriptions and object annotations available in the Visual Genome dataset, and using these to automatically construct attention maps that can be used for attention supervision, instead of requiring human annotators to manually provide grounding labels.
Our framework achieves competitive state-of-the-art VQA performance, while generating visual groundings that outperform other algorithms that use human annotated attention during training.


The contributions of this paper are: 
(1) we introduce a mechanism to automatically obtain meaningful attention supervision from both region descriptions and object annotations in the Visual Genome dataset;
(2) we show that by using the prediction of region and object label attention maps as auxiliary tasks in a VQA application, it is possible to obtain more interpretable intermediate representations.   
(3) we experimentally demonstrate state-of-the-art performances in VQA benchmarks as well as visual grounding that closely matches human attention annotations.


\section{Related Work}

Since its introduction \cite{Geman:EtAl:2015,VQAFirst:EtAl:2014,vqa}, the VQA 
problem has attracted an increasing interest \cite{VQA:Survey:2017}. Its multimodal 
nature and more precise evaluation protocol than alternative multimodal scenarios, such as image 
captioning, help to explain this interest. Furthermore, the proliferation of suitable 
datasets and potential applications, are also key elements behind this increasing activity. Most state-of-the-art methods follow a joint embedding approach, 
where deep models are used to project the textual question and visual input to a joint feature space 
that is then used to build the answer. Furthermore, most modern approaches pose VQA as a 
classification problem, where classes correspond to a set of pre-defined candidate 
answers. As an example, most entries to 
the VQA challenge \cite{vqa} select as output classes the most common 3000 answers in this 
dataset, which account for 92\% of the instances in the validation set. 

The strategy to combine the 
textual and visual embeddings and the underlying structure of the deep model are key design aspects that differentiate previous works.
Antol et al. \cite{vqa} propose an element-wise multiplication between image and 
question embeddings to generate spatial attention map. Fukui et al. \cite{mcb} propose multimodal 
compact bilinear pooling (MCB) to efficiently implement an outer product operator that 
combines visual and textual representations. Yu et al. \cite{mfh} extend this pooling scheme by introducing a multi-modal factorized bilinear pooling approach (MFB) that improves the representational capacity of the bilinear operator. They achieve this by adding an initial step that efficiently expands the textual and visual embeddings to a high-dimensional space. In terms of structural innovations, Noh et al. \cite{Noh:EtAl:2017} embed the textual question as an intermediate dynamic bilinear layer of a ConvNet that processes the visual information. Andreas et al. \cite{TreborGroup:EtAl:2016} propose a model that learns a set of task-specific neural modules that are jointly trained to answer visual questions. 

Following the successful introduction of soft attention in neural 
machine translation applications \cite{Bahdanau:EtAl:2015}, most modern VQA methods also incorporate a 
similar mechanism. The common approach is to use a one-way attention scheme, where the embedding of 
the question is used to generate a set of attention coefficients over a set of predefined image 
regions. These coefficients are then used to weight the embedding of the image regions to obtain 
a suitable descriptor \cite{Shih:EtAl:2016, winner:VQA:2017, mcb, mfb, mfh}. More 
elaborated forms of attention has also been proposed. Xu and Saenko \cite{xu2016ask} suggest use word-level embedding to generate attention. Yang et al. \cite{StackedAttention:CVPR:2016} 
iterates the application of a soft-attention mechanism over the visual input as a way to 
progressively refine the location of relevant cues to answer the question. Lu et al. 
\cite{CoattentionVQA:NIPS:2016} proposes a bidirectional co-attention mechanism that
besides the question guided visual attention, also incorporates a visual guided attention over the input question. 

In all the previous cases, the attention mechanism is applied using an unsupervised scheme, where 
attention coefficients are considered as latent variables. Recently, there have been also interest 
on including a supervised attention scheme to the VQA problem \cite{vqa-hat,Gan:EtAl:2017,VQA-Supervised:AAAI:2018}. Das et 
al. \cite{vqa-hat} compare the image areas selected by humans and state-of-the-art VQA techniques to 
answer the same visual question. To achieve this, they collect the VQA human attention 
dataset  (VQA-HAT), a large dataset of human attention maps built by asking humans to select images areas relevant to answer questions from the VQA dataset \cite{vqa}. Interestingly, this study 
concludes that current machine-generated attention maps exhibit a poor correlation with respect to 
the human counterpart, suggesting that humans use different visual cues to answer the questions. At a more fundamental level, this suggests that the discriminative nature of most current VQA 
systems does not effectively constraint the attention modules, leading to the encoding of discriminative cues instead of the underlying semantic that relates a given question-answer pair. Our findings in this work  support this hypothesis.

Related to the work in \cite{vqa-hat}, Gan et al. \cite{Gan:EtAl:2017} apply a more 
structured approach to identify the image areas used by humans to answer visual questions. 
For VQA pairs associated to images in the COCO dataset, they ask humans to select the 
segmented areas in COCO images that are relevant to answer each question. Afterwards, they use these 
areas as labels to train a deep learning model that is able to identify attention features. By 
augmenting a standard VQA technique with these attention features, they are able to achieve a small 
boost in performance. Closely related to our approach, Qiao et al. 
\cite{VQA-Supervised:AAAI:2018} use the attention labels in the VQA-HAT dataset to train an 
attention proposal network that is able to predict image areas relevant to answer a visual question. This network generates a set of attention proposals for each image in the VQA dataset, which are used as labels to supervise attention in the VQA model.
This strategy results in a small boost in performance compared with a non-attentional strategy. In contrast to our approach, these previous works are based on a supervised attention scheme that does not consider an automatic mechanism to obtain the attention labels. Instead, they rely on human annotated groundings as attention supervision. Furthermore, they differ from our work in the method to integrate attention labels to a VQA model.

\section{VQA Model Structure}\label{sec:vqa-model}
\begin{figure*}[tb]
\begin{center}
\centerline{\includegraphics[width=0.95\textwidth]{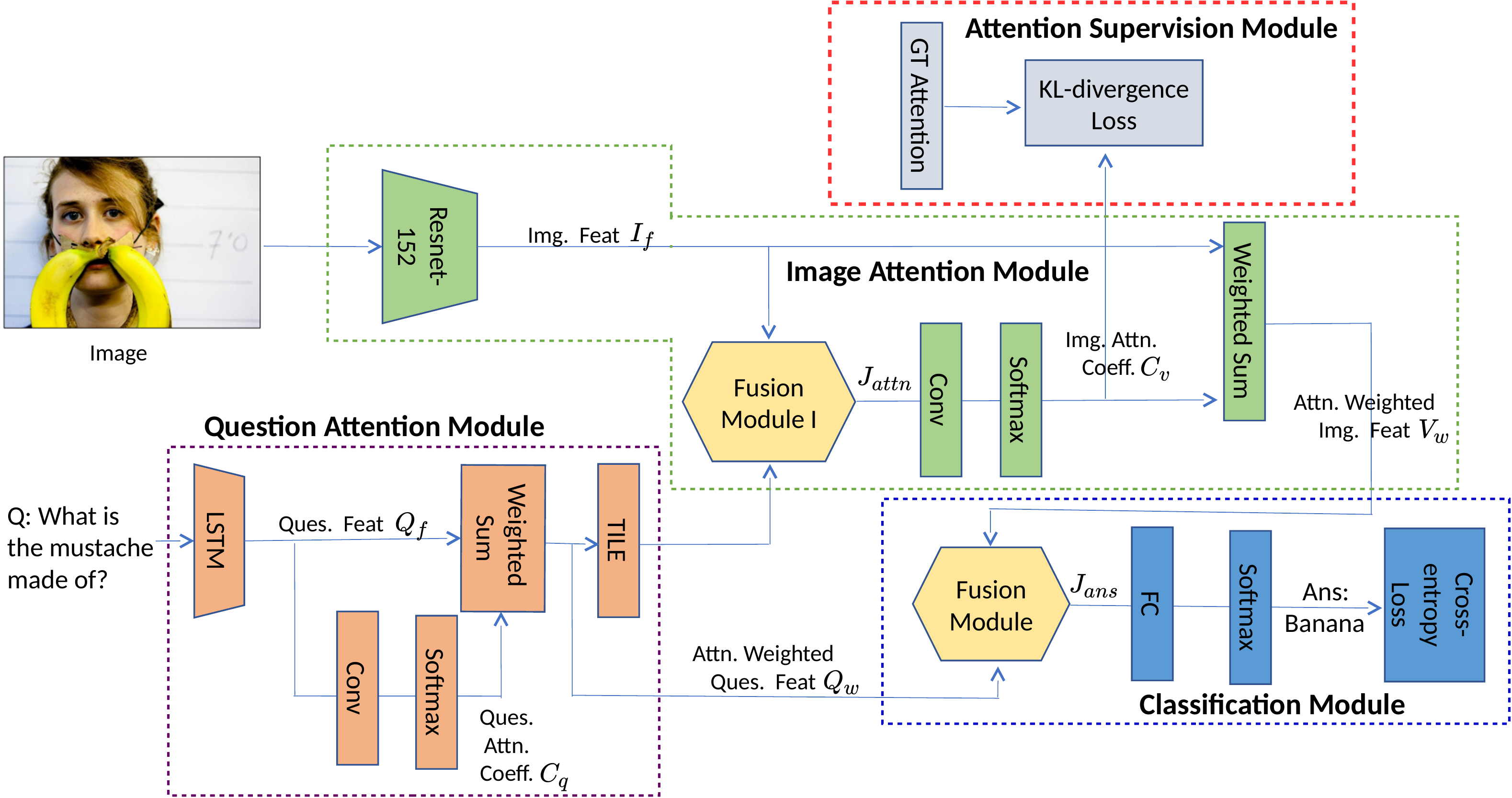}}
\caption{Schematic diagram of the main parts of the VQA model. It is mostly based on the model 
presented in \cite{mcb}. Main innovation is the Attention Supervision Module that incorporates visual grounding as an auxiliary task. This module is trained through the use of a set of image 
attention labels that are automatically mined from the Visual Genome dataset.}
\label{fig:model}
\end{center}
\end{figure*}

Figure \ref{fig:model} shows the main pipeline of our VQA model. We mostly build upon the MCB model 
in \cite{mcb}, which exemplifies current state-of-the-art 
techniques for this problem. Our main innovation to this model is the addition of 
an Attention Supervision Module that incorporates visual grounding as an auxiliary task. Next we describe the main modules behind this model.

\noindent \textbf{Question Attention Module:} Questions are tokenized and passed through an embedding layer, followed by an LSTM layer that 
generates the question features $Q_f\in \mathbb{R}^{T\times D}$, where $T$ is the maximum number of 
words in the tokenized version of the question and $D$ is the dimensionality of the hidden 
state of the LSTM. Additionally, following \cite{mfb}, a question 
attention mechanism is added that generates question attention coefficients 
$C_q\in\mathbb{R}^{T\times G_q}$, where $G_q$ is the so-called number of ``glimpses''. The purpose of 
$G_q$ is to allow the model to predict multiple attention maps so as to increase its expressiveness. 
Here, we use $G_q=2$. The weighted question features $Q_w\in \mathbb{R}^{G_{q}D}$ 
are then computed using a soft attention mechanism \cite{Bahdanau:EtAl:2015}, which is essentially 
a weighted sum of the $T$ word features followed by a concatenation according to $G_q$.

\noindent \textbf{Image Attention Module:} Images are passed through an embedding layer consisting of a pre-trained ConvNet model, such as 
Resnet pretrained with the ImageNet dataset \cite{he15deepresidual}. This generates image features 
$I_f\in \mathbb{R}^{C\times H\times W}$, where $C$, $H$ and $W$ are depth, height, and width of the 
extracted feature maps. Fusion Module I is then used to generate a set of image attention 
coefficients. First, question features $Q_w$ are tiled as the same spatial shape of $I_f$. 
Afterwards, the fusion module models the joint relationship $J_{attn}\in\mathbb{R}^{O\times H\times 
W}$ between questions and images, mapping them to a common space $\mathbb{R}^{O}$. In the simplest case, one can 
implement the fusion module using either concatenation or Hadamard product \cite{rcnn}, but more 
effective pooling schemes can be applied \cite{mcb,mlb,mfb,mfh}. The design choice of the fusion 
module remains an on-going research topic. In general, it should both effectively capture the latent 
relationship between multi-modal features meanwhile be easy to optimize. The fusion results are then 
passed through an attention module that computes the visual attention coefficient $C_v \in 
\mathbb{R}^{H\times W\times G_v}$, with which we can obtain attention-weighted visual features 
$V_w\in\mathbb{R}^{G_{v}C}$. Again, $G_v$ is the number of ``glimpses'', where we use $G_v=2$.

\noindent \textbf{Classification Module:} Using the compact representation of questions 
$Q_w$ and visual information $V_w$, the classification module applies first the Fusion Module II 
that provides the feature representation of answers $J_{ans}\in\mathbb{R}^{L}$, where 
$L$ is the latent answer space. Afterwards, it computes the logits over a 
set of predefined candidate answers. Following previous work \cite{mcb}, we use as candidate outputs the top 3000 
most frequent answers in the VQA dataset. At the end of this process, we obtain the highest scoring answer $\hat{A}$.

\noindent \textbf{Attention Supervision Module:} As a main novelty of the VQA model, we add an Image Attention Supervision Module as an auxiliary classification task, where ground-truth visual grounding labels $C_{gt}\in\mathbb{R}^{H\times W\times G_v}$ are used to guide the model to focus on meaningful parts of the image to answer each question. To do that, we simply treat the generated attention coefficients $C_v$ as a probability distribution, and then compare it with the ground-truth using KL-divergence. Interestingly, we introduce two attention maps, corresponding to relevant region-level and object-level groundings, as shown in Figure \ref{fig:region-ground}. Sections \ref{sec:vg_mine} and \ref{sec:implementation} provide details about our proposed method to obtain the attention labels and to train the resulting model, respectively. 

\section{Mining Attention Supervision from Visual Genome}\label{sec:vg_mine}
\begin{figure*}[tb]
\begin{center}
\begin{tabular}{cc}
\multicolumn{2}{c}{\textbf{(a)} Region-level grounding.} 
\\
\multicolumn{2}{c}{\small \textbf{Q:} What are the people doing? \textbf{Ans:} Talking.}
\\
\includegraphics[width=0.35\textwidth]{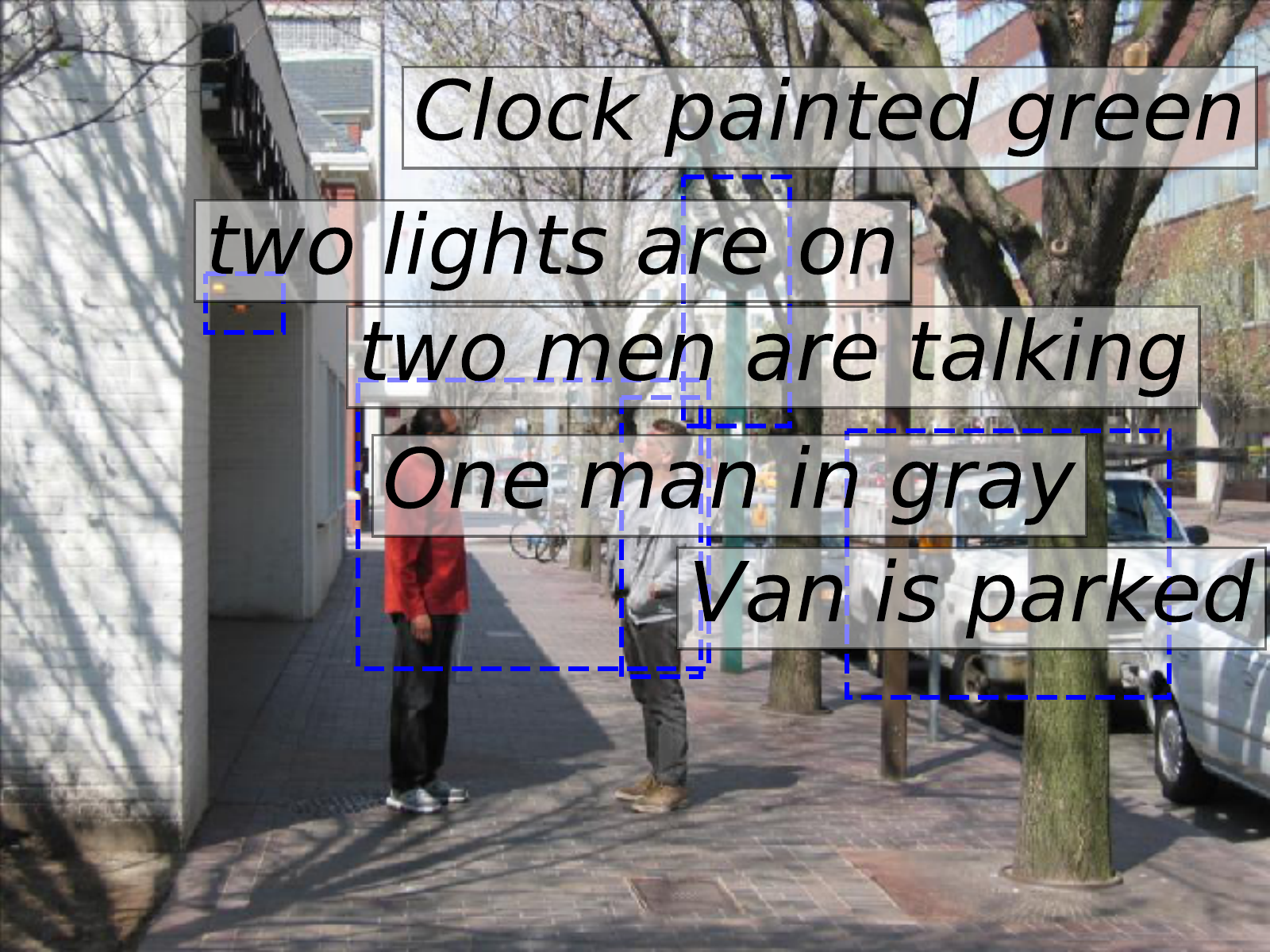}
\hspace{0.3cm}
& \includegraphics[width=0.35\textwidth]{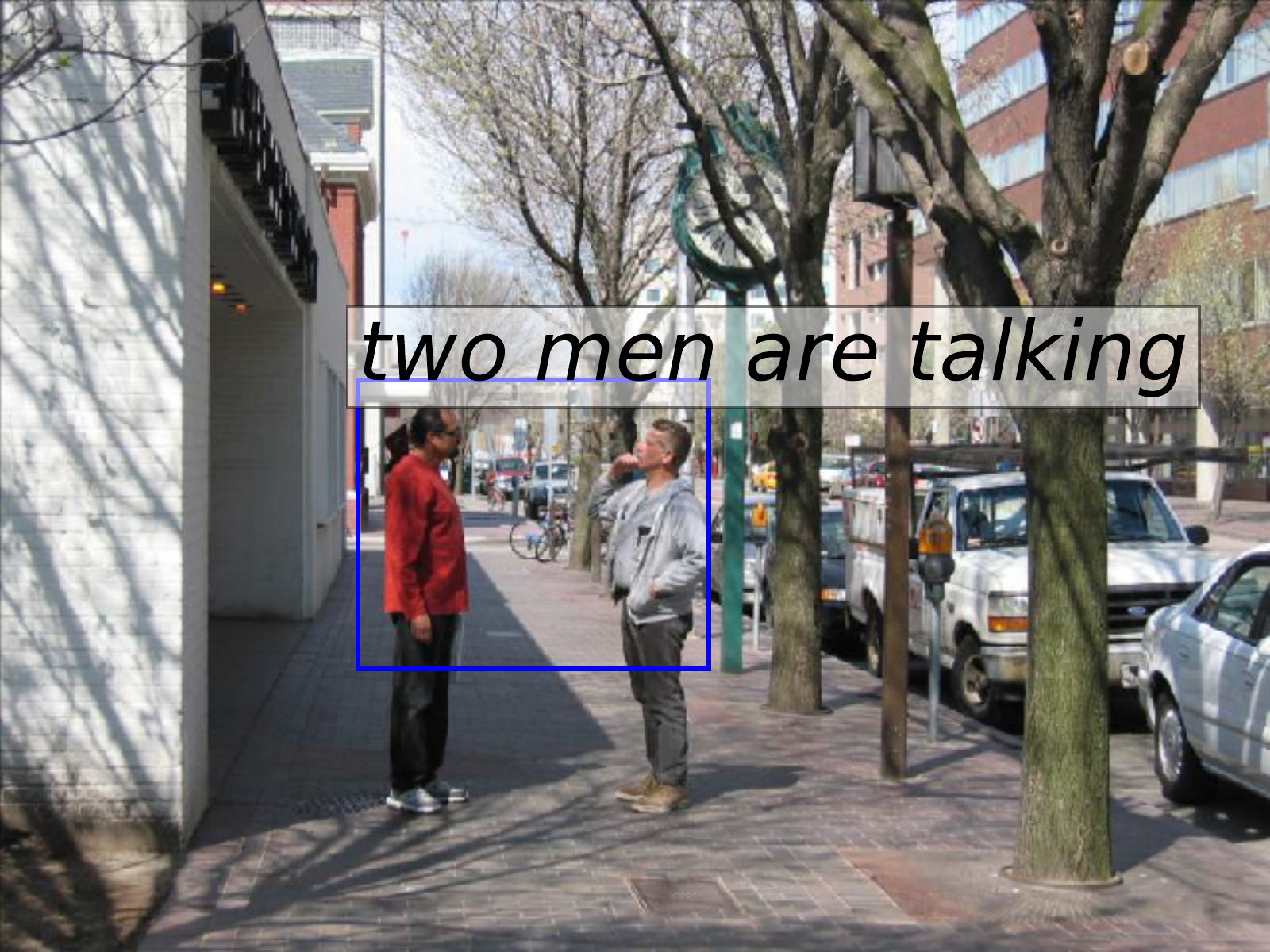}
\\
\\
\multicolumn{2}{c}{\textbf{(b)} Object-level grounding.} 
\\
\multicolumn{2}{c}{\small \textbf{Q:} How many people are there?  \textbf{Ans:} Two.}
\\
\includegraphics[width=0.35\textwidth]{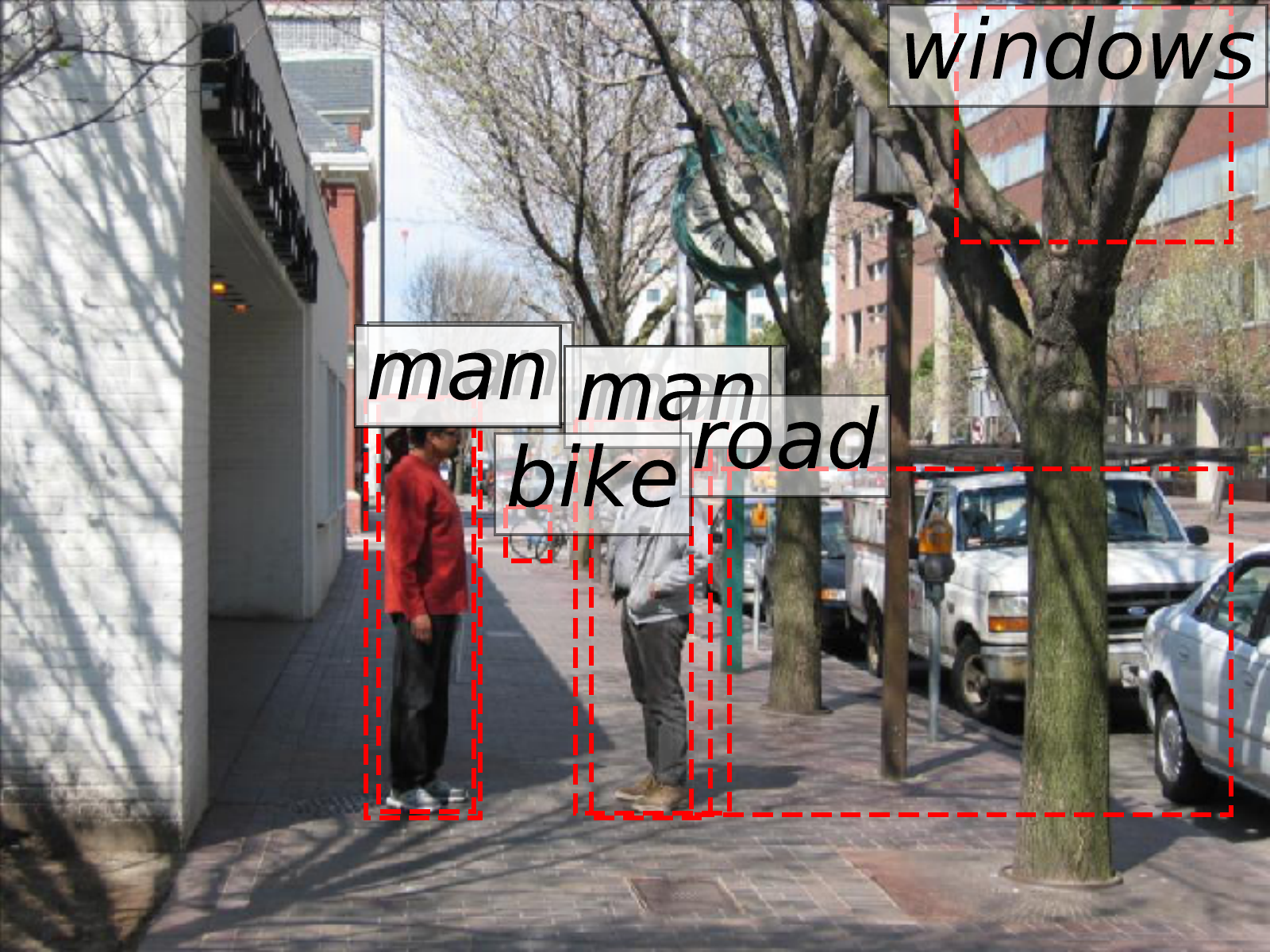}
\hspace{0.3cm}
& \includegraphics[width=0.35\textwidth]{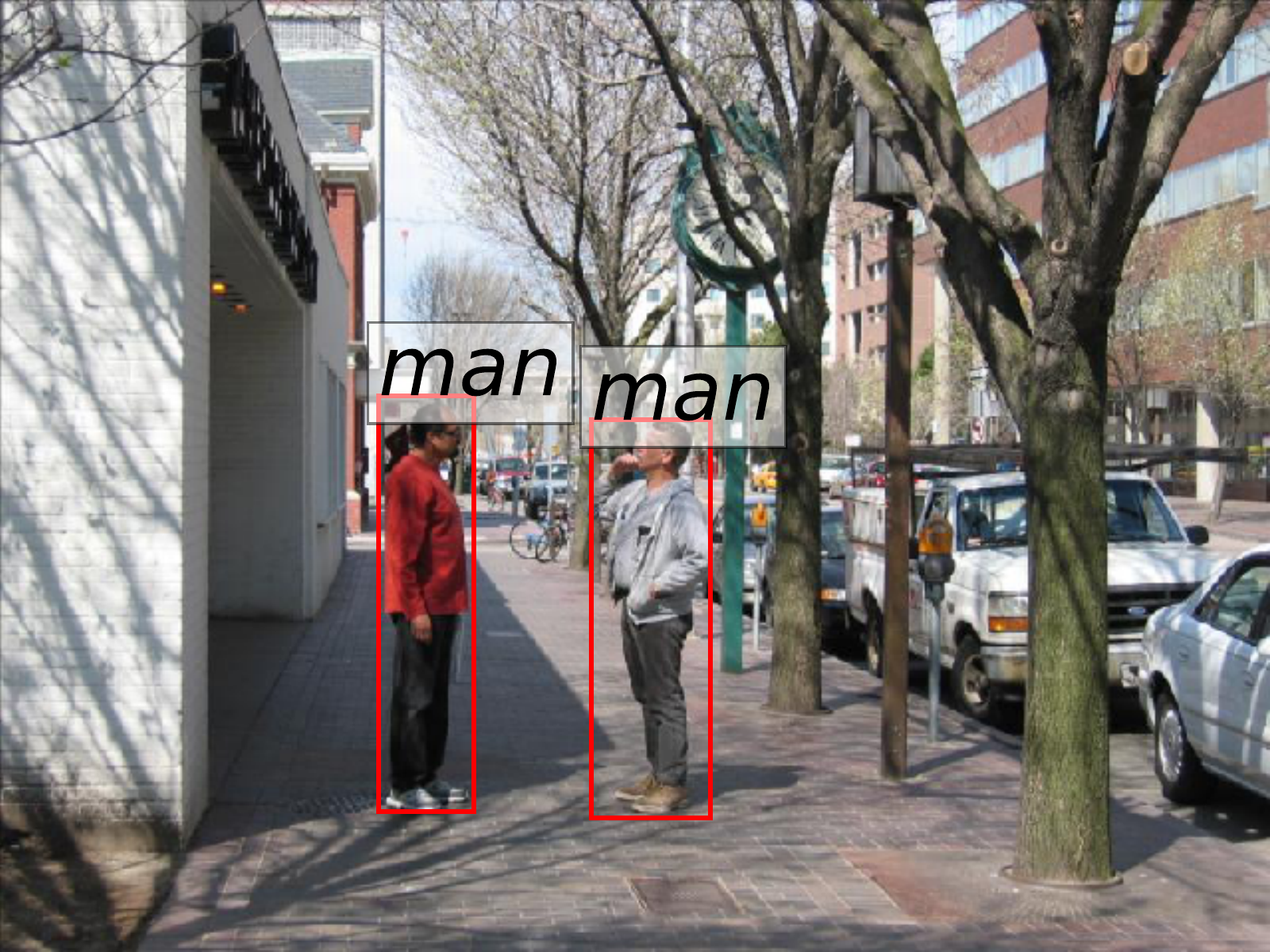}
\end{tabular}
\caption{\textbf{(a)} Example region-level groundings from VG. Left: image with region description labels; Right: our mined results. Here ``men'' in the region description is firstly lemmatized to be ``man'', whose aliases contain ``people''; the word ``talking'' in the answer also contributes to the matching. So the selected regions have two matchings which is the most among all candidates. 
\textbf{(b)} Example object-level grounding from VG. Left: image with object instance labels; Right: our mined results. Note that in this case region-level grounding will give us the same result as in (a), but object-level grounding is clearly more localized.}
\label{fig:region-ground}
\end{center}
\end{figure*}

Visual Genome (VG) \cite{Krishna2017} includes the largest VQA dataset currently available, which consists of 1.7M QA pairs. Furthermore, for each of its more than 100K images, VG also provides 
region and object annotations by means of bounding boxes. In terms of visual grounding, 
these region and object annotations provide complementary information. As an example, as shown in 
Figure \ref{fig:region-ground}, for questions related to interaction between 
objects, region annotations result highly relevant. In contrast, for questions related 
to properties of specific objects, object annotations result more valuable. Consequently, in this 
section we present a method to automatically select region and object annotations from VG that can 
be used as labels to implement visual grounding as an auxiliary task for VQA. 

For region annotations, we propose a simple heuristic to mine visual groundings:
for each $(I, Q, A)$ we 
enumerate all the region descriptions of $I$ and pick the description $D_i$ that has the 
most (at least two) overlapped informative words with $Q$ and $A$. Informative words 
are all nouns and verbs, where two informative words are matched if at least one of the 
following conditions is met:
(1) Their raw text as they appear in $Q$ or $A$ are the same;
(2) Their lemmatizations (using NLTK \cite{bird2004nltk}) are the same;
(3) Their synsets in WordNet \cite{miller1995wordnet} are the same;
(4) Their aliases (provided from VG) are the same. We refer to the resulting labels as \textit{region-level} groundings. Figure 
\ref{fig:region-ground}(a) illustrates an example of a region-level grounding.


In terms of object annotations, for each image in a $(I, Q, A)$ triplet we select the bounding
box of an object as a valid grounding label, if the object name matches one of the
informative nouns in $Q$ or $A$. To score each match, we use the same criteria as 
region-level groundings. Additionally, if a triplet $(I, Q, A)$ has a valid region grounding, each corresponding 
object-level grounding must be inside this region to be accepted as valid. As a further refinement, selected 
objects grounding are passed through an intersection over union filter to account for the 
fact that VG usually includes multiple labels for the same object instance. As a final consideration, for questions related to counting, region-level groundings are discarded after the corresponding object-level groundings are extracted. We refer to the resulting labels as \textit{object-level} groundings. Figure 
\ref{fig:region-ground}(b) illustrates an example of an object-level grounding.

As a result, combining both region-level and object-level groundings, about 700K out of 1M $(I, Q, A)$ triplets in VG end up with valid grounding labels. We will make these labels publicly available.

\section{Implementation Details}\label{sec:implementation}
We build the attention supervision on top of the open-sourced implementation of MCB \cite{mcb} and MFB \cite{mfb}. Similar to them,  We extract the image feature from \textit{res5c} layer of Resnet-152, resulting in $14\times 14$ spatial grid ($H=14$, $W=14$, $C=2048$). We construct our ground-truth visual grounding labels to be $G_v=2$ glimpse maps per QA pair, where the first map is object-level grounding and the second map is region-level grounding, as discussed in Section \ref{sec:vg_mine}. Let $(x^i_{min}, y^i_{min}, x^i_{max}, y^i_{max})$ be the coordinate of $i^{th}$ selected object bounding box in the grounding labels, then the mined object-level attention maps $C_{gt}^{0}$ are:
{\small \begin{equation}
C_{gt}^{0}[x, y] = \sum_{i \in objects}\mathcal{I}[x^i_{min}\le  x\le x^i_{max}] \; \mathcal{I}[y^i_{min}\le  y\le y^i_{max}]
\end{equation}
}where $\mathcal{I}[\cdot]$ is the indicator function.
Similarly, the region-level attention maps $C_{gt}^{1}$ are:
{\small
\begin{equation}
C_{gt}^{1}[x, y] = \sum_{i \in regions}\mathcal{I}[x^i_{min}\le  x\le x^i_{max}] \; \mathcal{I}[y^i_{min}\le  y\le y^i_{max}]
\end{equation}
}

Afterwards, $C_{gt}^{0}$ and $C_{gt}^{1}$ are spatially L1-normalized to represent probabilities and concatenated to form $C_{gt}\in \mathbb{R}^{14\times 14\times 2}$.

The model is trained using a multi-task loss,
{\small
\begin{equation}\label{eq:loss}
\begin{split}
    L(A, C_v, C_{gt}, \hat{A}|I, Q;\Theta) = & CE(A, \hat{A}|I, Q;\Theta) \\
    & + \alpha(t)KL(C_{gt}, C_v|I, Q;\Theta),
\end{split}
\end{equation}
}where $CE$ denotes cross-entropy and $KL$ denotes KL-divergence. $\Theta$ corresponds to the learned parameters. $\alpha(t)$ is a scalar that weights the loss terms. This scalar decays as a function of the iteration number $t$. In particular, we choose to use a cosine-decay function:
{\small
\begin{equation}\label{eq:alpha}
\alpha(t)=0.5\left(1+\cos(\pi\frac{t}{t_{max}})\right).
\end{equation}
}This is motivated by the fact that the visual grounding labels have some level of subjectivity. As an example, Figure \ref{fig:compare} (second row) shows a case where the learned attention seems more accurate than the VQA-HAT ground truth. Hence, as the model learns suitable parameter values, we gradually loose the penalty on the attention maps to provide more freedom to the model to selectively decide what attention to use. It is important to note that, 
for training samples in VQA-2.0 or VG that do not have region-level or object-level grounding labels, $\alpha = 0$
in Equation \ref{eq:loss}, so the loss is reduced to the classification term only. 
In our experiment, $t_{max}$ is calibrated for each tested model based on the number of training steps. In particular, we choose $t_{max}=190K$ for all MCB models and $t_{max}=160K$ for others.

\section{Experiments}
\subsection{Datasets}
\textbf{VQA-2.0:}
The VQA-2.0 dataset \cite{vqa} consists of 204721 images, with a total of 1.1M questions and 10 crowd-sourced answers per question. There are more than 20 question types, covering a variety of topics and free-form answers. The dataset is split into training (82K images and 443K questions), validation (40K images and 214K questions), and testing (81K images and 448K questions) sets. The task is to predict a correct answer $A$ given a corresponding image-question pair $(I, Q)$. As a main advantage with respect to version 1.0 \cite{vqa}, for every question VQA-2.0 includes complementary images that lead to different answers, reducing language bias by forcing the model to use the visual information.

\noindent \textbf{Visual Genome:}
The Visual Genome (VG) dataset \cite{Krishna2017} contains 108077 images, with an average of 17 QA pairs per image. We follow the processing scheme from \cite{mcb}, where non-informative words in the questions and answers such as ``a'' and ``is'' are removed. Afterwards, $(I, Q, A)$ triplets with answers to be single keyword and overlapped with VQA-2.0 dataset are included in our training set. This adds 97697 images and about 1 million questions to the training set. 
Besides the VQA data, VG also provides on average 50 region descriptions and 30 object instances per image. Each region/object is annotated by one sentence/phrase description and bounding box coordinates. 

\noindent \textbf{VQA-HAT:}
VQA-HAT dataset \cite{vqa-hat} contains 58475 human visual attention heat (HAT) maps for $(I, Q, A)$ triplets in VQA-1.0 training set. Annotators were shown a blurred image, a $(Q, A)$ pair and were asked to ``scratch'' the image until they believe someone else can answer the question by looking at the blurred image and the sharpened area. The authors also collect $1374\times 3=4122$ HAT maps for VQA-1.0 validation sets, where each of the 1374 $(I, Q, A)$ were labeled by three different annotators, so one can compare the level of agreement among labels. We use VQA-HAT to evaluate visual grounding performance, by comparing the rank-correlation between human attention and model attention, as in \cite{vqa-hat,vqa-x}.

\noindent \textbf{VQA-X:}
VQA-X dataset \cite{vqa-x} contains 2000 labeled attention maps in VQA-2.0 validation sets. In contrast to VQA-HAT, VQA-X attention maps are in the form of instance segmentations, where annotators were asked to segment objects and/or regions that most prominently justify the answer. Hence the attentions are more specific and localized. We use VQA-X to evaluate visual grounding performance by comparing the rank-correlation, as in \cite{vqa-hat,vqa-x}.

\subsection{Results}

\begin{table}[t]
\small
\centering
\begin{tabular}{|lccc|}
\hline
 & \multicolumn{2}{c}{{Rank Correlation}} & Accuracy/\%\\
 & VQA-HAT & VQA-X & VQA-2.0\\\hline
Human \cite{vqa-hat} & 0.623 & - & 80.62 \\ 
\hline
PJ-X \cite{vqa-x} & 0.396 & 0.342 & - \\
MCB \cite{mcb} &  0.276 & 0.261 & 62.27\\
Attn-MCB, $\alpha$=1 (ours) &  \textbf{0.580} & \textbf{0.396} & 60.51\\
Attn-MCB (ours) &  0.517 & 0.375 & 62.24\\
MFB \cite{mfb} & 0.276 & 0.299 & 65.22\\
Attn-MFB (ours) &  0.416 & 0.335 & 65.36\\
MFH \cite{mfh} &  0.354 & 0.350 & 66.17\\
Attn-MFH (ours) &  0.483 & 0.376 & \textbf{66.31}\\
\hline
\end{tabular}
\caption{\label{tab:vanilla} Evaluation of different VQA models on visual grounding and answer prediction. The reported accuracies are evaluated using the VQA-2.0 test-standard set.}
\end{table}

\begin{figure*}[h!]
\begin{center}
\begin{tabular}{ccc}
{\small VQA-HAT Ground Truth} & {\small MFH} & {\small Attn-MFH (Ours)} \\
\hline
\\
\multicolumn{3}{c}{\textbf{Q:} Is the computer on or off? \textbf{Ans:} on}\\
\includegraphics[width=0.3\textwidth]{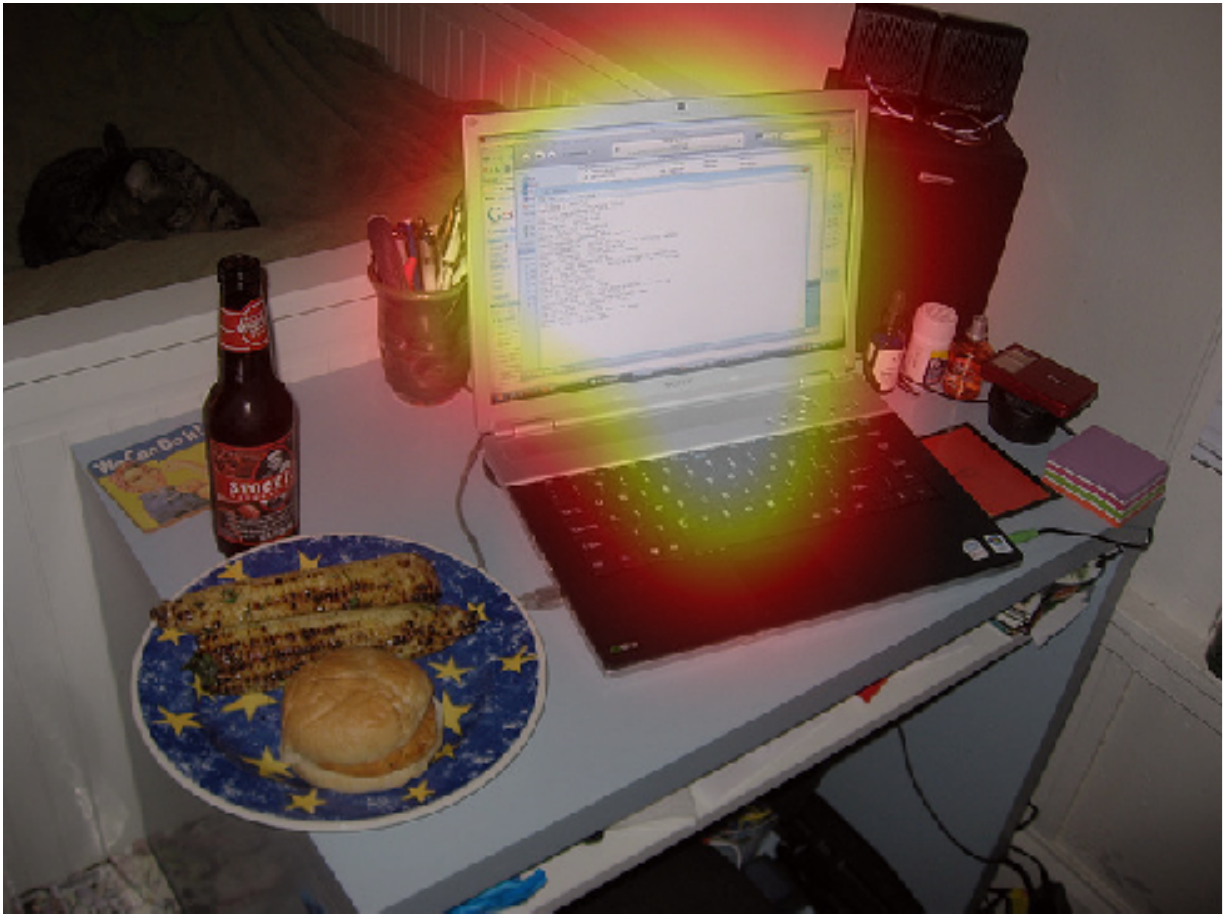}
& \includegraphics[width=0.3\textwidth]{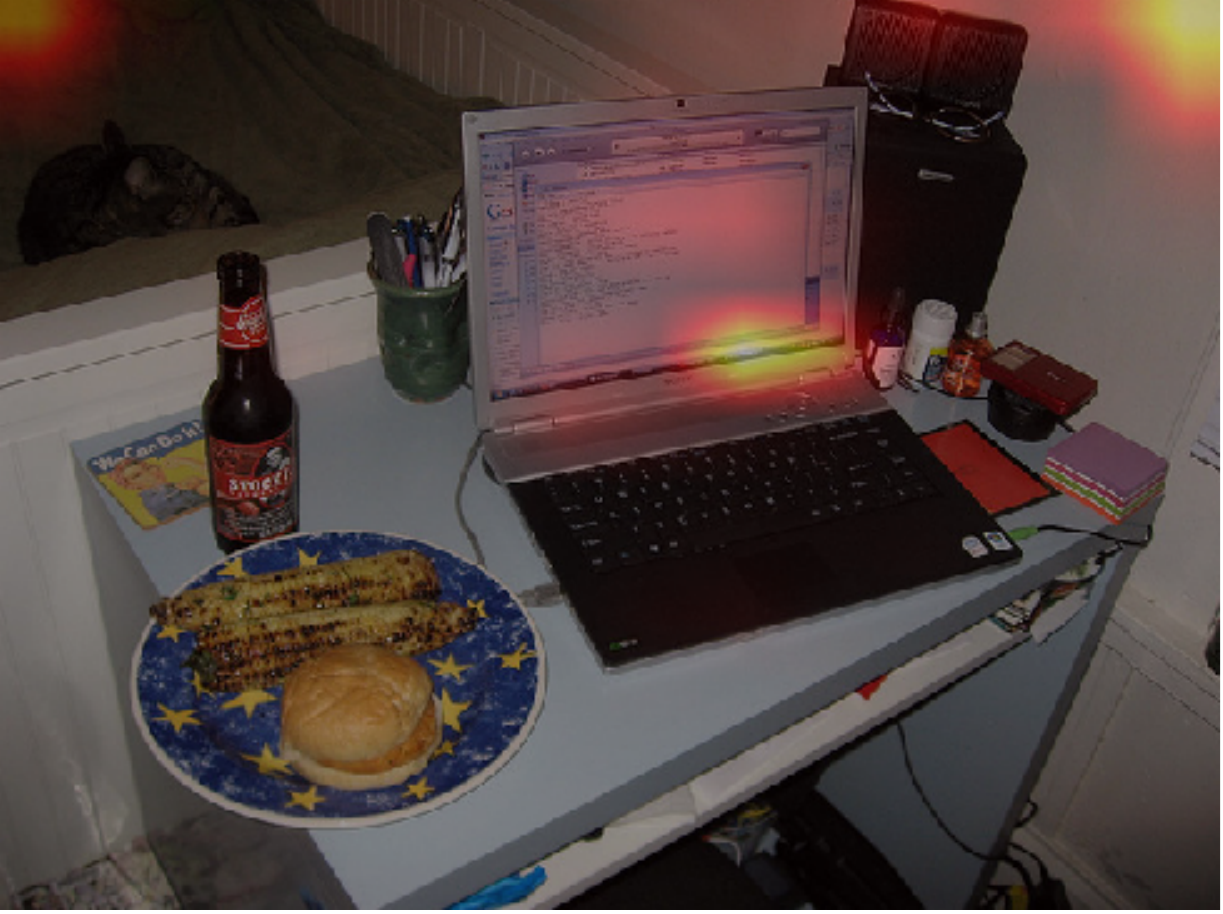}
& \includegraphics[width=0.3\textwidth]{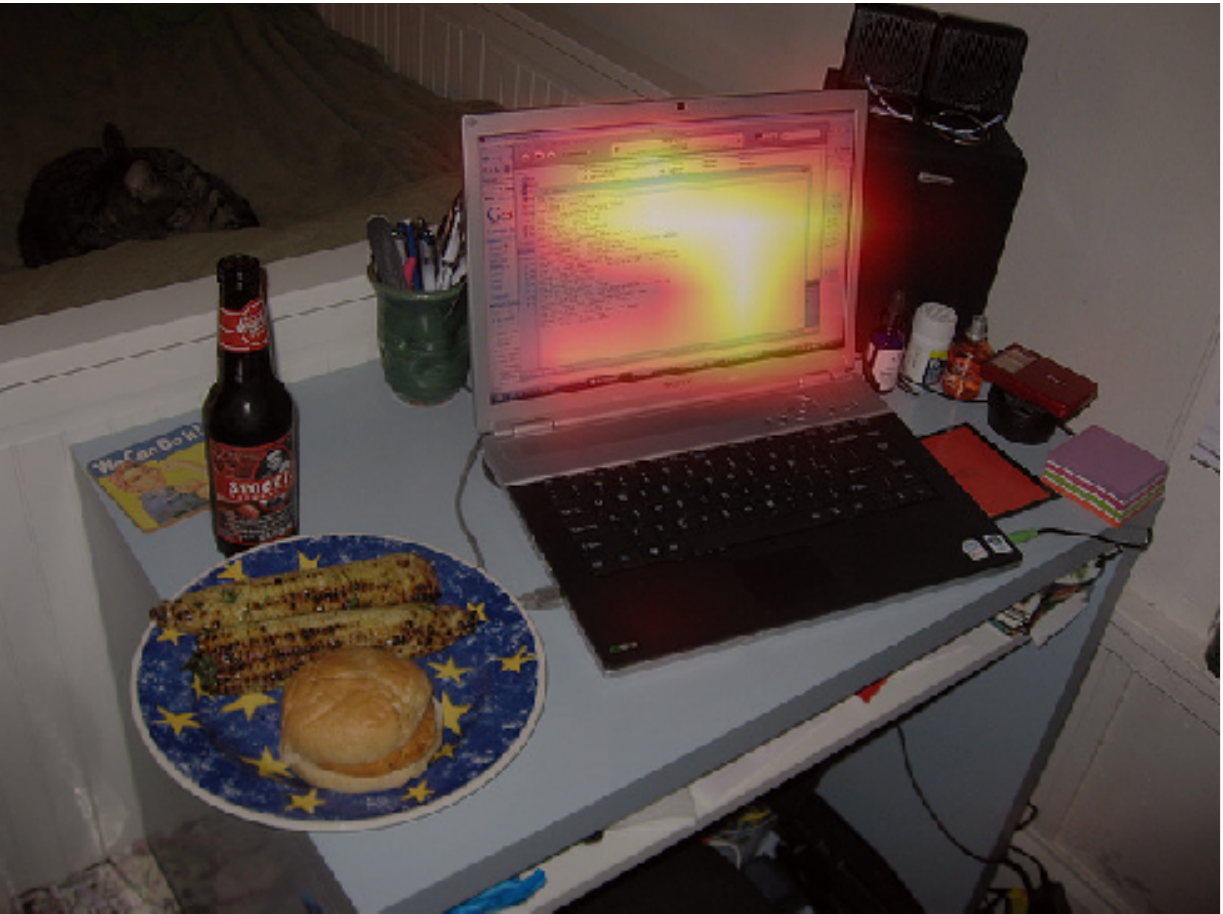} \\
\\
\multicolumn{3}{c}{\textbf{Q:} What color is the inside of the cats ears? \textbf{Ans:} pink}\\
\includegraphics[width=0.3\textwidth]{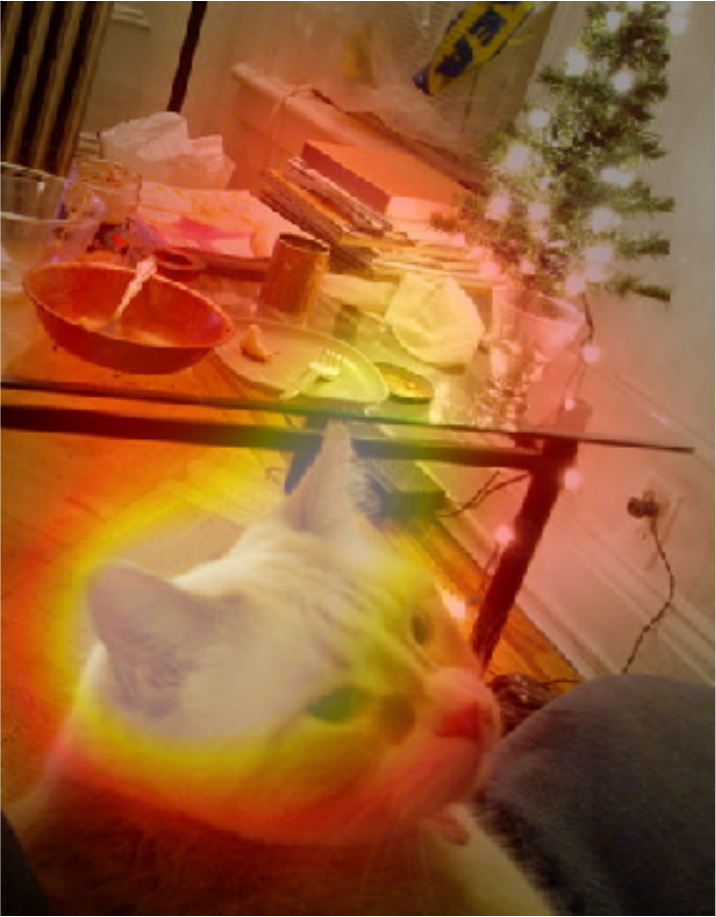}
& \includegraphics[width=0.3\textwidth]{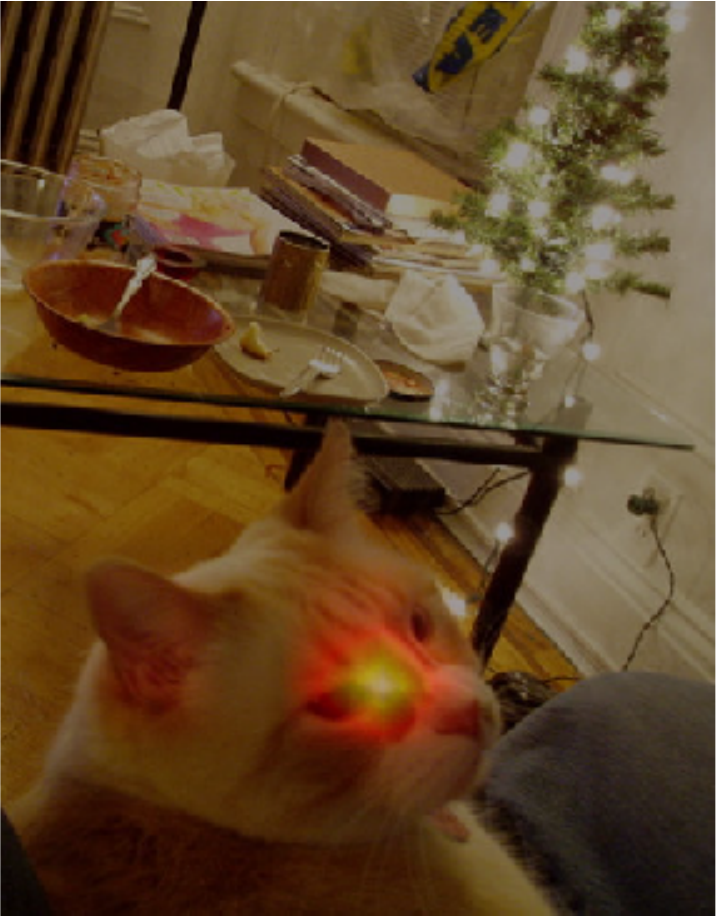}
& \includegraphics[width=0.3\textwidth]{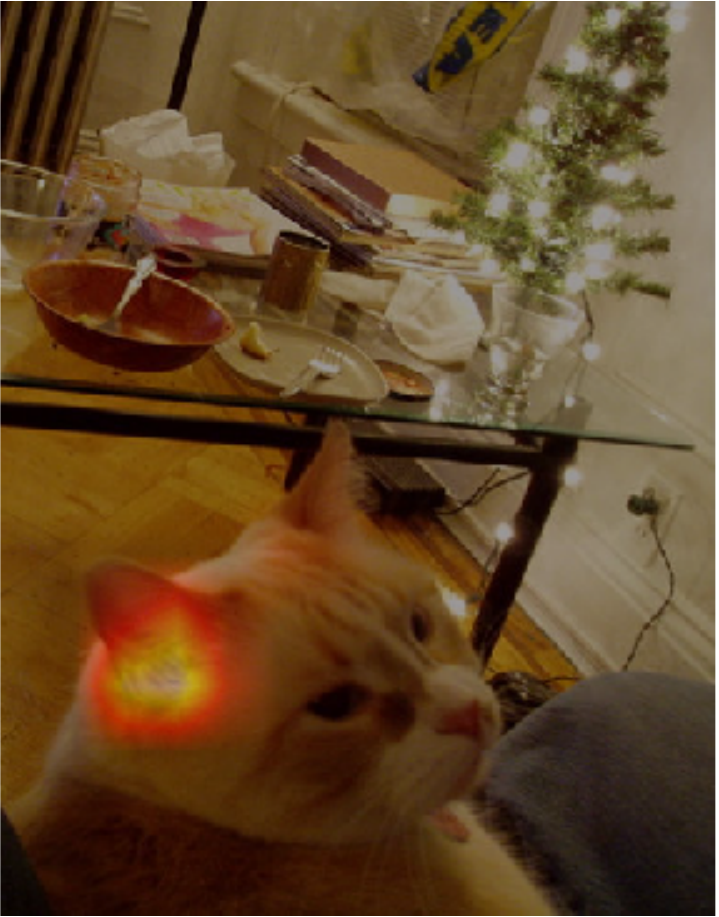} \\
\\
\multicolumn{3}{c}{\textbf{Q:} How many of these animals are there? \textbf{Ans:} 2}\\
\includegraphics[width=0.3\textwidth]{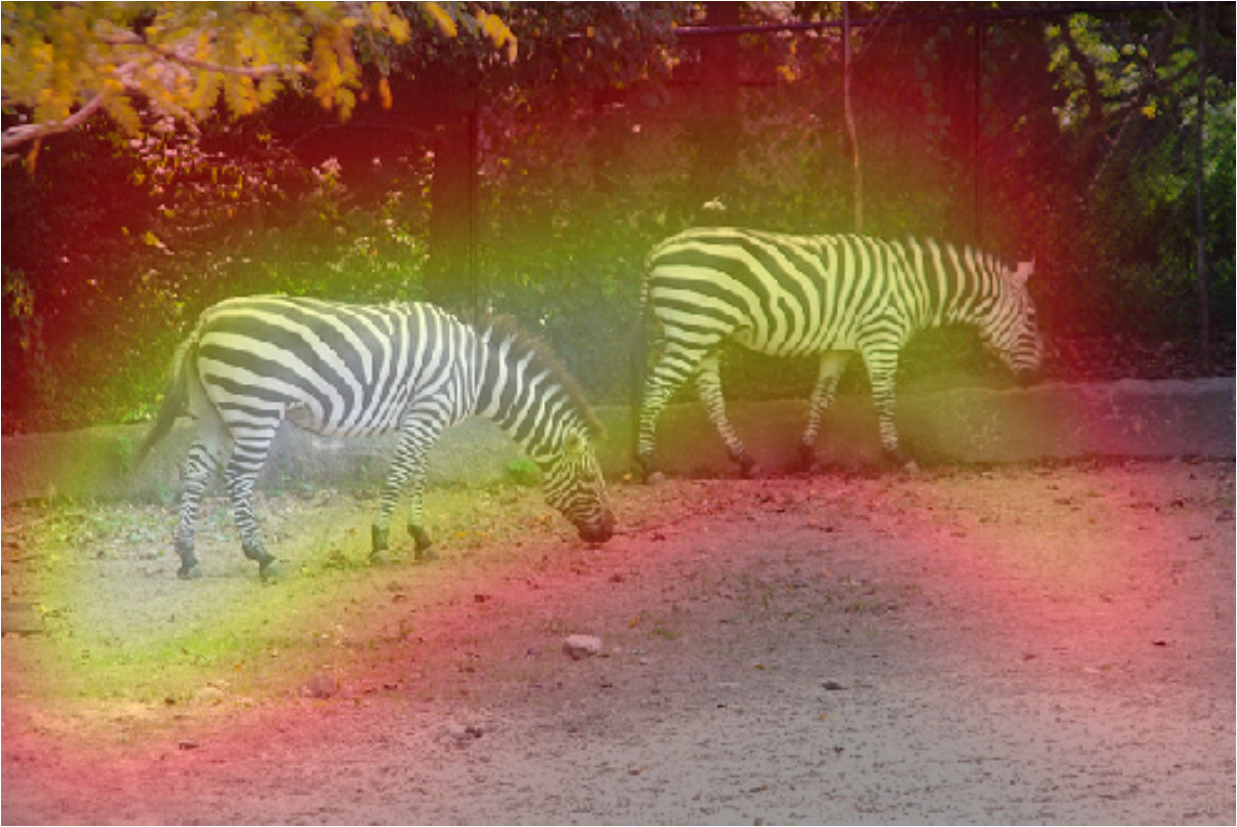}
& \includegraphics[width=0.3\textwidth]{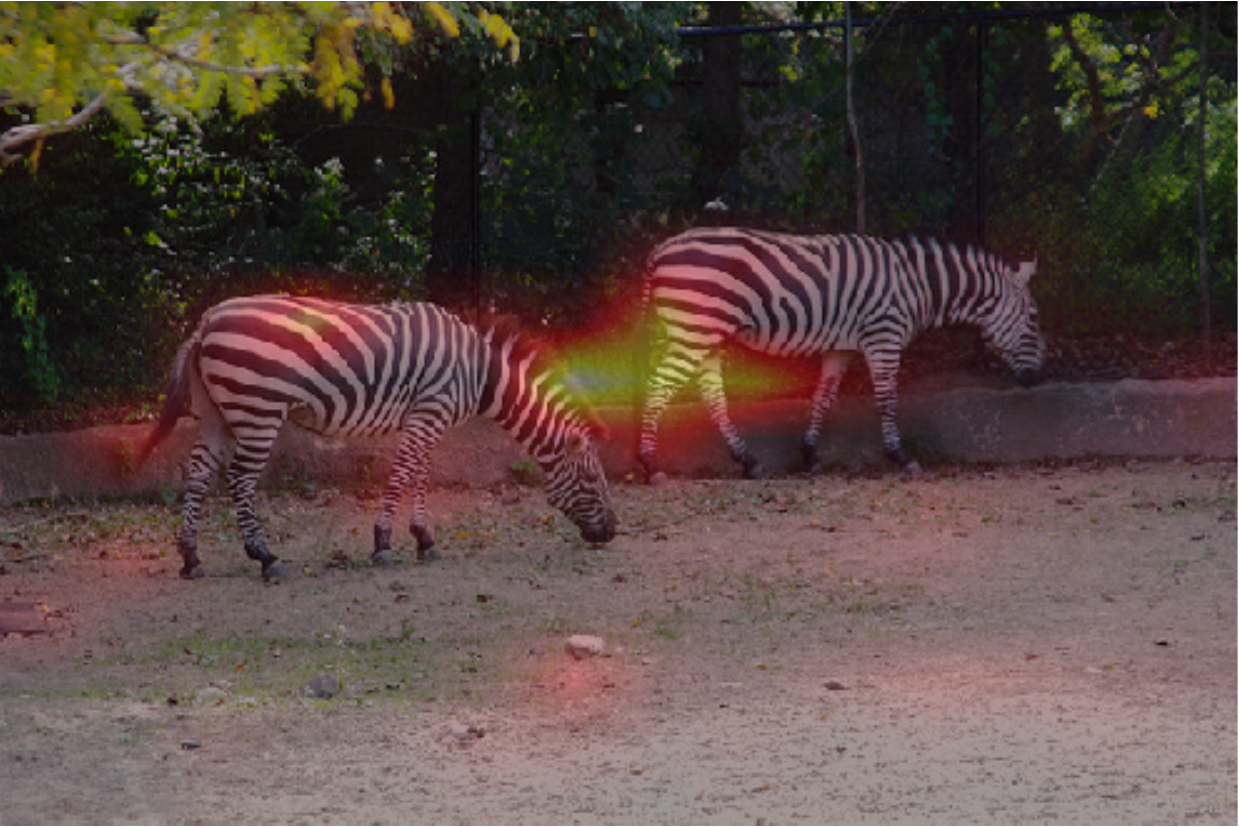}
& \includegraphics[width=0.3\textwidth]{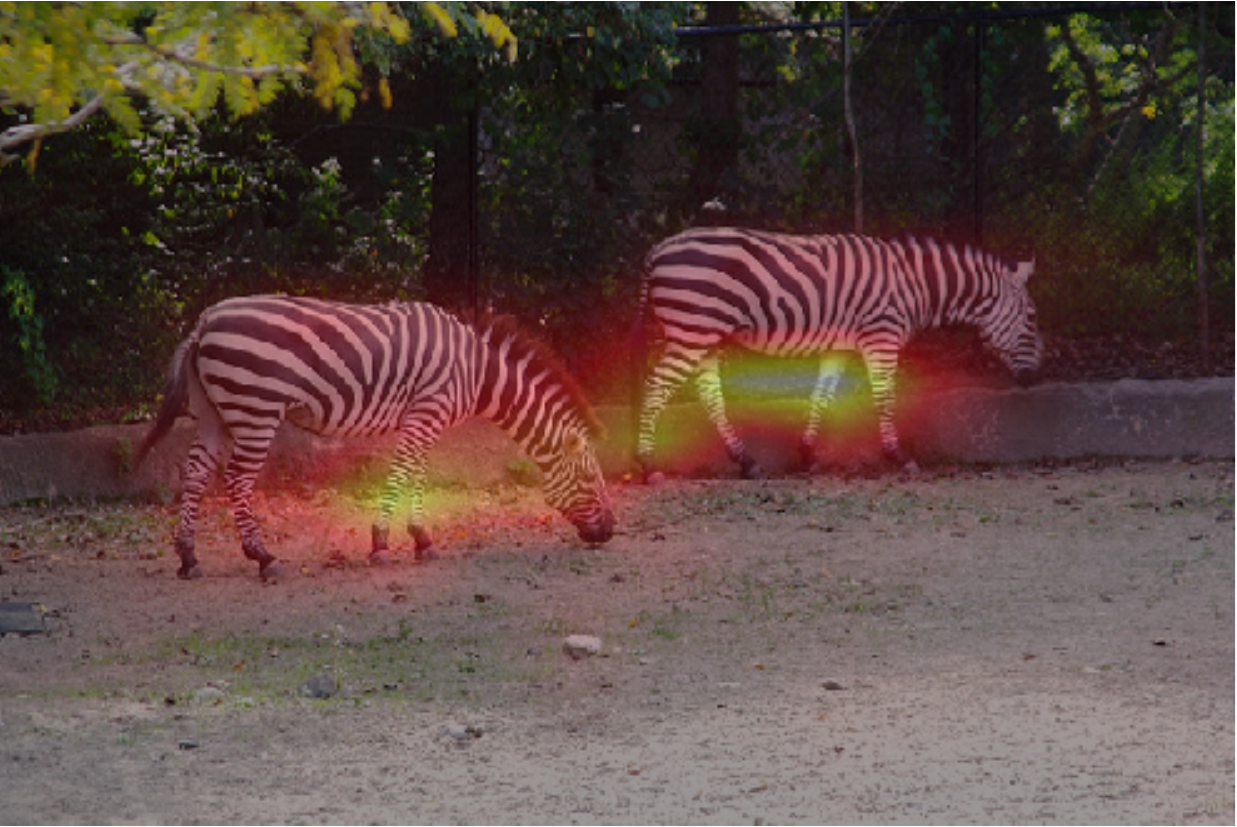}
\end{tabular}
\caption{Visual grounding comparison: the first column is the ground-truth human attention in VQA-HAT \cite{vqa-hat}; the second column shows the results from pretrained MFH model \cite{mfh}; the last column are our Attn-MFH trained with attention supervision. We can see that the attention areas considered by our model mimic the attention areas used by humans, but they are more localized in space.}
\label{fig:compare}
\end{center}
\end{figure*}

We evaluate the performance of our proposed method using two criteria: i) rank-correlation \cite{spearman1904proof} to evaluate visual grounding and ii) accuracy to evaluate question answering. Intuitively, rank-correlation measures the similarity between human and model attention maps under a rank-based metric. A high rank-correlation means that the model is `looking at' image areas that agree to the visual information used by a human to answer the same question. In terms of accuracy of a predicted answer $\hat{A}$ is evaluated by:
{\small
\begin{equation}\label{eq:acc}
\textup{Accuracy}(\hat{A}) = \min\left\{\frac{\sum_{i=1}^{10}\mathcal{I}[\hat{A}=A_i]}{3}, 1\right\}.
\end{equation}
}

Table \ref{tab:vanilla} reports our main results. Our models are built on top of prior works with the additional Attention Supervision Module as described in Section \ref{sec:vqa-model}. Specifically, we denote by Attn-* our adaptation of the respective model by including our Attention Supervision Module. We  highlight that MCB model is the winner of VQA challenge 2016 and MFH model is the best single model in VQA challenge 2017. In Table \ref{tab:vanilla}, we can observe that our proposed model achieves a significantly boost on rank-correlation with respect to human attention. Furthermore, our model outperforms alternative state-of-art techniques in terms of accuracy in answer prediction. Specifically, the rank-correlation for MFH model increases by 36.4\% when is evaluated in VQA-HAT dataset and 7.7\% when is evaluated in VQA-X. This indicates that our proposed methods enable VQA models to provide more meaningful and interpretable results by generating more accurate visual grounding. 

Table \ref{tab:vanilla} also reports the result of an experiment where the decaying factor $\alpha(t)$ in Equation \ref{eq:alpha} is fixed to a value of 1. In this case, the model is able to achieve higher rank-correlation, but accuracy drops by 2\%.  We observe that as training proceeds, attention loss becomes dominant in the final training steps, which affects the accuracy of the classification module. 

Figure \ref{fig:compare} shows qualitative results of the resulting visual grounding, including also a comparison with respect to no-attn model. 


\section{Conclusions}

In this work we have proposed a new method that is able to slightly outperform current state-of-the-art VQA systems, while also providing interpretable representations in the form of an explicitly trainable visual attention mechanism. Specifically, as a main result, our experiments provide evidence that the generated visual groundings achieve high correlation with respect to human-provided attention annotations, outperforming the correlation scores of previous works by a large margin. 

As further contributions, we highlight two relevant insides of the proposed approach. On one side, by using attention labels as an auxiliary task, the proposed approach demonstrates that is able to constraint the internal representation of the model in such a way that it fosters the encoding of interpretable representations of the underlying relations between the textual question and input image. On other side, the proposed approach demonstrates a method to leverage existing datasets with region descriptions and object labels to effectively supervise the attention mechanism in VQA applications, avoiding costly human labeling. 

As future work, we believe that the superior visual grounding provided by the proposed method can play a relevant role to generate natural language explanations to justify the answer to a given visual question. This scenario will help to demonstrate the relevance of our technique as a tool to increase the capabilities of AI based technologies to explain their decisions.

\vspace{0.2cm}

\textbf{Acknowledgements:} This work was partially funded by Oppo, Panasonic and the Millennium Institute for Foundational Research on Data.

{\small
\bibliographystyle{ieee}
\bibliography{egbib}
}

\end{document}